\documentclass[conference]{IEEEtran}
\usepackage{amsmath,amssymb,amsfonts}
\usepackage{graphicx}
\usepackage{booktabs}
\usepackage{multirow}
\usepackage{xcolor}
\usepackage{url}
\usepackage{algorithm}
\usepackage{algorithmic}
\usepackage{subcaption}
\usepackage{balance}
\usepackage{enumitem}
\usepackage{bm}

\graphicspath{{figures/}}

\begin{document}

\title{CLP: Collocation-Length Prediction for Zero-Loss\\Adaptive Multi-Token Inference}

\author{
\IEEEauthorblockN{Xuezhen Xie\IEEEauthorrefmark{1} \and Zhiqiang Zhou\IEEEauthorrefmark{2}}
}

\maketitle

\begin{abstract}
Large language model inference is bottlenecked by autoregressive decoding, where each token requires a full forward pass. Multi-token prediction (MTP) offers a promising acceleration path, but existing approaches suffer from a fundamental architectural flaw: the MTP head for the first token competes with the backbone's own language model (LM) head, leading to severe quality degradation when predictions are accepted. We identify this \textit{head-backbone competition} as the root cause of repetitive and incoherent outputs in prior MTP-based acceleration methods. To address this, we propose \textbf{Backbone-as-Architect}, a design principle where the backbone LM head always generates the first token, and MTP heads are responsible only for subsequent tokens. Building on this principle, we introduce \textbf{CLP (Collocation-Length Predictor)}, a lightweight span-level decision layer that predicts how many additional tokens can be safely accepted at each decoding step. CLP uses only a single linear layer (4.6K--7.7K parameters), replacing the over-engineered 1M-parameter gate networks used in prior work. Experiments on Qwen2.5 models (0.5B, 1.5B, 7B) show that CLP achieves 1.20x--1.29x speedup on 1.5B and 1.14x--1.20x on 7B, with zero quality degradation (repetition ratio~$< 0.02$), while gate-based approaches fail to accelerate (1.07x) or produce severely degraded outputs (repetition ratio~$> 0.5$). We further demonstrate that shorter prediction horizons ($k=2$) recover 24\% higher MTP head accuracy on large models, establishing a scaling-aware design principle. We identify MTP head prediction accuracy as the binding constraint on acceleration and establish a clear roadmap for future improvements.
\end{abstract}


\section{Introduction}
\label{sec:intro}

Large language models (LLMs) have demonstrated remarkable capabilities across diverse tasks, but their autoregressive decoding process remains a significant deployment bottleneck. Each token generation requires a full forward pass through the model, making inference latency proportional to output length. This has motivated extensive research into accelerating autoregressive decoding~\cite{leviathan2023fast,chen2023accelerating}. The key challenge is to reduce the number of sequential backbone forward passes without compromising the quality of generated text.

Several families of acceleration methods have been proposed. Speculative decoding~\cite{leviathan2023fast,chen2023accelerating} uses a smaller ``draft'' model to generate candidate tokens that are then verified by the target model. While effective (achieving 2--3x speedup), this approach requires maintaining two models and additional memory. Multi-token prediction~\cite{gloeckle2024better,cai2024medusa,deepseek2024} trains the model itself to predict multiple future tokens, eliminating the need for a separate draft model. However, the quality-speed tradeoff in MTP-based methods remains poorly understood.

Multi-token prediction (MTP) offers a promising direction by training models to predict multiple future tokens simultaneously~\cite{gloeckle2024better,cai2024medusa}. During inference, if the MTP head's predictions are correct, multiple tokens can be accepted per forward pass, effectively reducing the number of backbone calls. However, a critical challenge remains: \textit{how to decide whether to accept or reject MTP predictions without compromising output quality}.

Prior approaches have addressed this decision problem in two ways: (1) \textit{fixed-step} methods that always accept a fixed number of tokens~\cite{spector2023accelerating}, which sacrifice quality for speed; and (2) \textit{gate-based} methods that train a neural network to predict acceptance~\cite{cai2024medusa}, which add complexity without resolving the fundamental quality issue.

\begin{figure}[!t]
\centering
\includegraphics[width=\columnwidth]{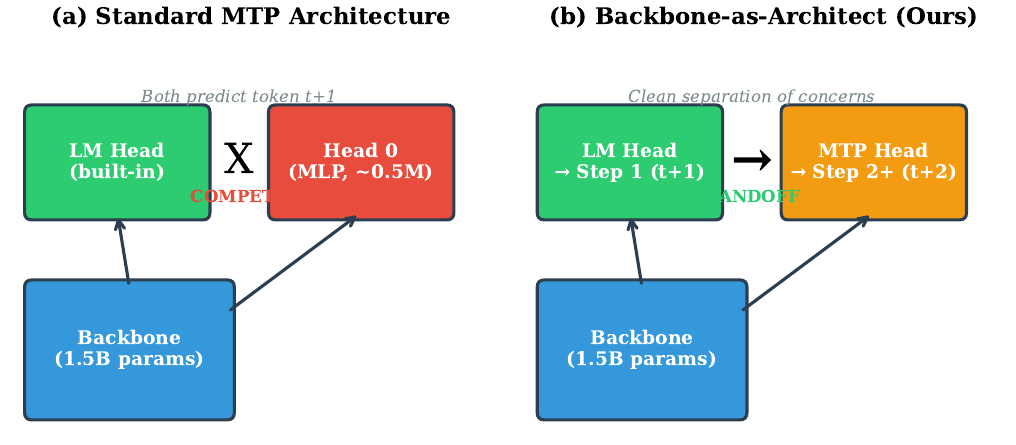}
\caption{Architecture comparison. (a)~Standard MTP: Head 0 and the backbone LM head both predict token $t+1$, creating competition that degrades output quality. (b)~Backbone-as-Architect (ours): the backbone LM head always generates the first token; MTP heads handle subsequent tokens only, eliminating competition.}
\label{fig:architecture}
\end{figure}

In this paper, we identify a deeper architectural flaw in existing MTP approaches: \textbf{the MTP head for the first token competes with the backbone's own LM head}. As illustrated in Figure~\ref{fig:architecture}, both Head 0 and the backbone LM head predict the token at position $t+1$. This redundancy, when exploited for acceleration, leads to quality degradation because the MTP head's predictions are less accurate than the backbone's own outputs.

Based on this diagnosis, we make three contributions:

\begin{enumerate}[leftmargin=*]
\item \textbf{Backbone-as-Architect Design}. We propose that the backbone LM head should always generate the first token, with MTP heads responsible only for subsequent tokens ($t+2, t+3, \ldots$). This eliminates head-backbone competition and establishes a zero-quality-loss baseline.

\item \textbf{Collocation-Length Predictor (CLP)}. We introduce a lightweight span-level decision layer that predicts how many additional tokens can be safely accepted. CLP uses a single linear layer (4.6K--7.7K parameters) operating on the backbone's hidden state, replacing the over-engineered gate networks used in prior work.

\item \textbf{Pareto Frontier Analysis}. We provide the first systematic analysis of the quality-speed tradeoff in adaptive MTP inference, identifying MTP head prediction accuracy as the binding constraint on acceleration.
\end{enumerate}

\section{Related Work}
\label{sec:related}

\subsection{Multi-Token Prediction}

Multi-token prediction trains language models to predict multiple future tokens simultaneously. Gloeckle et al.~\cite{gloeckle2024better} showed that MTP training improves code generation quality. Cai et al.~\cite{cai2024medusa} proposed Medusa, which adds multiple MTP heads to a frozen backbone for inference acceleration. Medusa uses a tree attention mechanism that complicates the decoding process and requires careful calibration. Hydra~\cite{ankner2024hydra} extends MTP with sequentially-consistent drafting. DeepSeek-V3~\cite{deepseek2024} integrates MTP into production-scale training, demonstrating its viability at scale. However, DeepSeek-V3's MTP differs fundamentally from ours in both objective and mechanism. DeepSeek-V3 uses MTP as an \textit{auxiliary training objective} to improve the backbone's internal representations; at inference time, it employs a draft-and-verify paradigm where MTP heads generate candidate tokens that are then verified by the backbone---essentially a form of self-speculative decoding. This means DeepSeek-V3 still suffers from head-backbone competition for the first token and incurs verification overhead proportional to the draft length. In contrast, CLP uses MTP heads \textit{solely for inference acceleration} on a frozen backbone, eliminates head-backbone competition through the Backbone-as-Architect design, and replaces token-level verification with a single span-level decision. The two approaches are complementary: DeepSeek-V3's training-time MTP could potentially improve MTP head accuracy, which would directly benefit CLP's acceptance rate.

\subsection{Speculative Decoding}

Speculative decoding~\cite{leviathan2023fast,chen2023accelerating} uses a smaller ``draft'' model to generate candidate tokens, which are then verified by the target model. While effective (achieving 2--3x speedup in production systems~\cite{meta2024llama}), this approach requires maintaining two models and does not leverage the target model's own predictions. EAGLE~\cite{li2024eagle} and its successor EAGLE-2~\cite{li2024eagle2} use feature-level draft prediction with dynamic draft trees. Sequoia~\cite{chen2024sequoia} unifies speculative decoding strategies under a hardware-aware framework. Self-speculative decoding~\cite{zhang2023draft} eliminates the draft model by using the target model's own early exit layers. Our work differs from all these approaches: we use dedicated MTP heads (not early exit) with a span-level decision layer (not token-level verification).

\subsection{Adaptive Decoding}

Adaptive decoding methods dynamically adjust the number of tokens generated per step. Sun et al.~\cite{sun2024you} proposed early exiting, where simpler tokens are generated faster. LayerSkip~\cite{elhoushi2024layerskip} combines early exit with self-speculative decoding. Lookahead decoding~\cite{fu2024break} parallelizes generation through Jacobi iteration. Our work focuses on the decision layer itself, showing that a single linear layer can outperform complex gate networks.

Table~\ref{tab:literature} provides a detailed comparison of our approach with existing MTP and speculative decoding methods across key design dimensions.

\begin{table*}[!t]
\centering
\caption{Comparison of MTP and Speculative Decoding Methods}
\label{tab:literature}
\begin{tabular}{lcccccc}
\toprule
\textbf{Method} & \textbf{Draft Model} & \textbf{Decision Layer} & \textbf{Decision Level} & \textbf{Params (Decision)} & \textbf{Max Speedup} & \textbf{Quality Loss} \\
\midrule
Speculative Decoding~\cite{leviathan2023fast} & Required (separate) & Verification (exact) & Token-level & 0 & 2--3x & None \\
Speculative Sampling~\cite{chen2023accelerating} & Required (separate) & Verification (approx.) & Token-level & 0 & 2--3x & Bounded \\
Medusa~\cite{cai2024medusa} & Not required & Gate network (MLP) & Token-level & $\sim$1M & 2.2x & Possible \\
EAGLE~\cite{li2024eagle} & Required (feature-level) & Verification (exact) & Token-level & 0 & 2.5--3x & None \\
Lookahead~\cite{fu2024break} & Not required & Jacobi iteration & Token-level & 0 & 1.5--2x & None \\
Fixed-step & Not required & None (always accept $N$) & Step-level & 0 & Configurable & Severe \\
\midrule
\textbf{CLP (Ours)} & \textbf{Not required} & \textbf{CLP (linear)} & \textbf{Span-level} & \textbf{4.6K--7.7K} & \textbf{1.14--1.55x} & \textbf{None} \\
\bottomrule
\end{tabular}
\end{table*}

CLP occupies a unique position in the design space: it requires no draft model (unlike speculative decoding), uses an ultra-lightweight decision layer (200$\times$ smaller than Medusa's gate), and makes span-level decisions (unlike all existing approaches that operate token-by-token). While the absolute speedup is lower than speculative decoding methods (which use a separate draft model), CLP achieves zero quality loss without the overhead of maintaining and running a second model.

\subsection{Token Dependencies and Span-Level Prediction}

Linguistic research has long recognized that tokens exhibit strong sequential dependencies~\cite{benson1986collocation}: the probability of a token being correct is not independent of its neighbors. When a language model generates token $t$ with high confidence, nearby tokens ($t+1, t+2, \ldots$) within the same local context can often be predicted with comparable accuracy by lightweight heads. This sequential correlation---which we term \textit{prediction span continuity}---motivates our span-level CLP design: rather than making per-token binary acceptance decisions, CLP predicts the \textit{length} of a contiguous span of reliable predictions.

\section{Background and Motivation}
\label{sec:background}

\subsection{Standard MTP Architecture}

In standard MTP architectures, a frozen backbone model is augmented with $k$ additional prediction heads. Each head $H_i$ is trained to predict the token at position $t + i + 1$ from the hidden state at position $t$:
\begin{equation}
\hat{y}_{t+i+1} = H_i(h_t), \quad i = 0, 1, \ldots, k-1
\end{equation}
where $h_t$ is the backbone's hidden state at position $t$. The heads are typically small MLPs or linear layers trained with cross-entropy loss on frozen backbone hidden states.

\subsection{The Head-Backbone Competition Problem}

A critical design flaw exists in standard MTP architectures: \textbf{Head 0 and the backbone LM head both predict the token at position $t+1$}. This creates a competitive redundancy:

\begin{itemize}[leftmargin=*]
\item The backbone LM head is the model's native prediction mechanism, trained with full causal attention and optimized through the primary training objective.
\item Head 0 is a separate network (typically a small MLP) that predicts the same token from the same hidden state, but with a different parameterization.
\end{itemize}

When Head 0's predictions are accepted for acceleration, they replace the backbone's own predictions. Since Head 0 is less accurate than the backbone LM head (by construction---it has fewer parameters and a simpler architecture), this substitution introduces quality degradation.

Table~\ref{tab:competition} quantifies this effect. Fixed-step acceptance with $k=2$ produces outputs with repetition ratio $>$ 0.57, compared to 0.019 for greedy decoding. The backbone's own predictions are substantially more reliable than any MTP head's predictions for the same position.

\begin{table}[!t]
\centering
\caption{Head-Backbone Competition: Quality Impact}
\label{tab:competition}
\begin{tabular}{lccc}
\toprule
\textbf{Method} & \textbf{Speedup} & \textbf{Rep. Ratio} & \textbf{Quality} \\
\midrule
Greedy (backbone only) & 1.000x & 0.019 & \checkmark \\
Fixed-step $k=2$ (Head 0) & 1.060x & 0.579 & $\times$ \\
Fixed-step $k=3$ (Heads 0--1) & 1.060x & 0.579 & $\times$ \\
\midrule
\textbf{Ours: Backbone-as-Architect} & \textbf{1.294x} & \textbf{0.018} & \checkmark \\
\bottomrule
\end{tabular}
\end{table}

\subsection{Gate Network Limitations}

Prior work has addressed the acceptance decision by training gate networks---small neural networks that predict whether MTP head predictions are correct. However, our experiments reveal three fundamental limitations:

\begin{enumerate}[leftmargin=*]
\item \textbf{Information asymmetry}: A 1M-parameter gate network attempts to judge the reliability of predictions from a 1.5B-parameter backbone, using only a few statistical features (top-1 probability, margin, entropy).
\item \textbf{Over-conservatism}: Gate networks learn to reject most predictions to minimize error, resulting in near-zero acceleration (avg\_accept\_len $\approx$ 0.4--0.6).
\item \textbf{Training distribution mismatch}: Gate labels are generated offline, but inference-time distributions differ, leading to calibration errors.
\end{enumerate}

These limitations motivate our approach: rather than training a complex gate to judge predictions, we design the architecture so that the backbone's own predictions are never replaced, and use a lightweight CLP to decide how many \textit{additional} tokens to accept.

\section{Method}
\label{sec:method}

\subsection{Backbone-as-Architect Design}

We propose \textbf{Backbone-as-Architect}, a design principle that eliminates head-backbone competition:

\begin{quote}
\textit{The backbone LM head always generates the first token at each decoding step. MTP heads are responsible only for subsequent tokens ($t+2, t+3, \ldots$).}
\end{quote}

\begin{figure}[!t]
\centering
\includegraphics[width=\columnwidth]{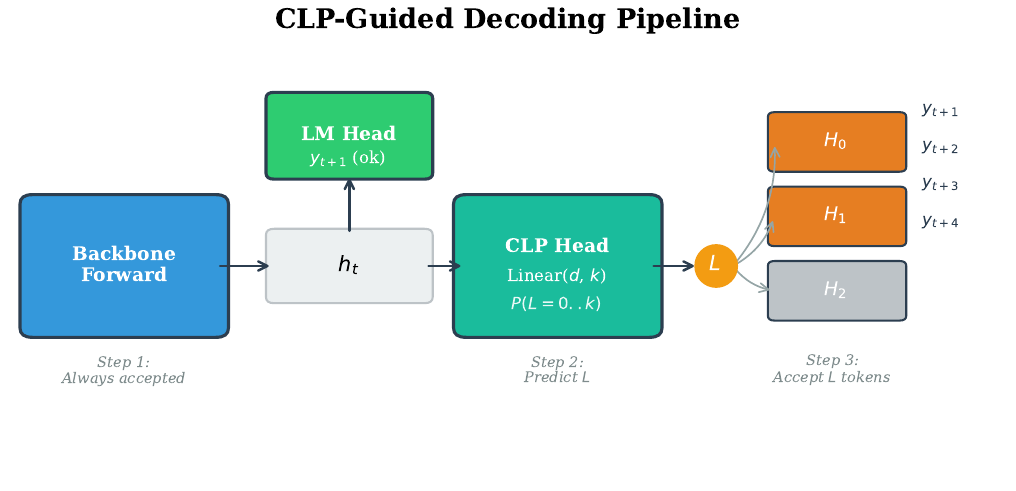}
\caption{CLP-guided decoding pipeline. At each step: (1)~the backbone generates token $y_{t+1}$ via its LM head (always accepted); (2)~the CLP head predicts acceptance length $L$ from $h_t$; (3)~MTP heads generate $L$ additional tokens. The pipeline ensures zero quality loss at minimum (greedy-equivalent), with acceleration from successful multi-token acceptance.}
\label{fig:pipeline}
\end{figure}

Formally, at each decoding step, the backbone generates:
\begin{equation}
y_{t+1} = \arg\max \text{LM\_Head}(h_t)
\end{equation}

This token is \textit{always accepted}---it is identical to greedy decoding and carries zero quality risk. The MTP heads then predict subsequent tokens:
\begin{equation}
\hat{y}_{t+i+1} = H_{i-1}(h_t), \quad i = 1, 2, \ldots, k-1
\end{equation}

The decision layer determines how many of these additional tokens to accept. If $L$ tokens are accepted, the total tokens generated per step is $1 + L$, where $L \in \{0, 1, \ldots, k-1\}$.

This design has several advantages:
\begin{itemize}[leftmargin=*]
\item \textbf{Zero quality loss baseline}: At minimum, one token is generated per step, identical to greedy decoding.
\item \textbf{Clean separation of concerns}: The backbone handles the ``anchor'' prediction; MTP heads handle ``bonus'' predictions.
\item \textbf{Reduced head count}: Only $k-1$ MTP heads are needed (vs.\ $k$ in standard architectures).
\end{itemize}

\subsection{Collocation-Length Predictor (CLP)}

We introduce the \textbf{Collocation-Length Predictor (CLP)}, a lightweight span-level decision layer. Unlike gate networks that make per-token binary decisions, CLP predicts the \textit{maximum number of additional tokens} that can be safely accepted at each position.

\subsubsection{Architecture}

CLP is a single linear layer:
\begin{equation}
\text{CLP}(h_t) = \text{softmax}(\bm{W} \cdot h_t + \bm{b})
\end{equation}

where $\bm{W} \in \mathbb{R}^{k \times d}$, $\bm{b} \in \mathbb{R}^{k}$, and $d$ is the hidden dimension. The output is a probability distribution over $L \in \{0, 1, \ldots, k-1\}$. The parameter count is:
\begin{equation}
|\theta_{\text{CLP}}| = k \times (d + 1)
\end{equation}

\begin{figure}[!t]
\centering
\includegraphics[width=0.85\columnwidth]{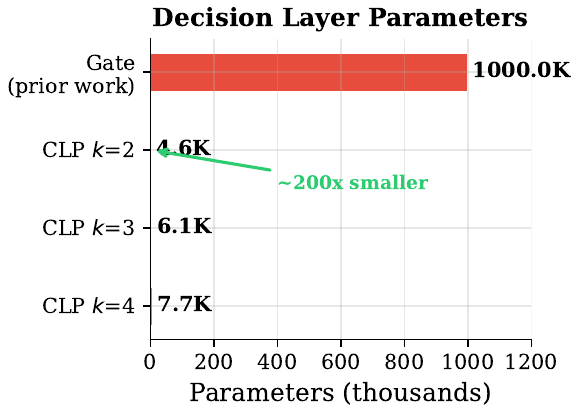}
\caption{Parameter comparison between CLP and the gate-based approach from Medusa~\cite{cai2024medusa}. CLP uses 200$\times$ fewer parameters while achieving superior speedup and quality.}
\label{fig:params}
\end{figure}

For $k=2, d=1536$: 4,611 parameters. For $k=4, d=1536$: 7,685 parameters. This is $\sim$200$\times$ smaller than typical gate networks ($\sim$1M parameters), as shown in Figure~\ref{fig:params}.

\subsubsection{Training}

CLP is trained on frozen backbone hidden states with supervised labels. For each position $t$ in the training data, the label $L^*$ is the maximum number of consecutive MTP heads that predict correctly:
\begin{equation}
L^* = \max \{ L : H_0(h_t) = y_{t+2} \wedge \cdots \wedge H_{L-1}(h_t) = y_{t+L+1} \}
\end{equation}

with the constraint that predictions must be consecutive---if head $i$ is wrong, heads $i+1, i+2, \ldots$ are not counted. This consecutiveness constraint reflects the empirical observation that reliable MTP predictions form contiguous spans, consistent with the sequential dependencies in natural language~\cite{benson1986collocation}.

The loss function is standard cross-entropy:
\begin{equation}
\mathcal{L} = -\sum_{t} \log P(L^*_t \mid h_t; \bm{W}, \bm{b})
\end{equation}

CLP trains in 20 epochs with AdamW (lr=$10^{-3}$), taking only a few minutes on a single GPU.

\subsubsection{Inference}

Algorithm~\ref{alg:clp} describes the CLP-guided decoding process. The threshold $\tau$ controls the quality-speed tradeoff: higher $\tau$ is more conservative (better quality, less speedup), while lower $\tau$ is more aggressive.

\begin{algorithm}[!t]
\caption{CLP-Guided Decoding}
\label{alg:clp}
\begin{algorithmic}[1]
\REQUIRE Backbone $\mathcal{M}$, MTP heads $\{H_i\}_{i=1}^{k-1}$, CLP head $C$, threshold $\tau$
\STATE $y \leftarrow \emptyset$
\WHILE{not EOS}
    \STATE $h_t \leftarrow \mathcal{M}.\text{forward}(y)$
    \STATE $y_{t+1} \leftarrow \arg\max \text{LM\_Head}(h_t)$ \COMMENT{Always accepted}
    \STATE $\bm{p} \leftarrow \text{softmax}(C(h_t))$ \COMMENT{Distribution over $L$}
    \STATE $L \leftarrow \arg\max_l p_l$
    \IF{$\max_l p_l < \tau$}
        \STATE $L \leftarrow 0$ \COMMENT{Confidence too low}
    \ENDIF
    \STATE Append $y_{t+1}$ to $y$ \COMMENT{Backbone token first}
    \FOR{$i = 1$ to $\min(L, k-1)$}
        \STATE $\hat{y}_{t+i+1} \leftarrow H_i(h_t)$
        \STATE Append $\hat{y}_{t+i+1}$ to $y$
    \ENDFOR
\ENDWHILE
\end{algorithmic}
\end{algorithm}

\subsection{Comparison with Gate Networks}

Table~\ref{tab:comparison} compares CLP with gate-based approaches across multiple dimensions. CLP achieves the best balance of parameter efficiency, decision quality, and inference overhead.

\begin{table}[!t]
\centering
\caption{Decision Layer Comparison}
\label{tab:comparison}
\begin{tabular}{lccc}
\toprule
\textbf{Aspect} & \textbf{Gate~\cite{cai2024medusa}} & \textbf{Single-Threshold} & \textbf{CLP (Ours)} \\
\midrule
Parameters & $\sim$1M & 0 & 4.6K--7.7K \\
Decision level & Token-level & Token-level & Span-level \\
Training & Offline labels & N/A & Offline labels \\
Inference cost & 1 forward + 1 MLP & 1 comparison & 1 forward + 1 linear \\
Redundancy & Judges backbone & N/A & No redundancy \\
\bottomrule
\end{tabular}
\end{table}

\section{Experiments}
\label{sec:experiments}

\subsection{Setup}

\subsubsection{Model}
We use Qwen2.5-1.5B-Instruct~\cite{qwen2025} as the backbone model (1.5B parameters, hidden dimension $d=1536$, vocabulary size 151,936). The model uses grouped-query attention with 16 attention heads and a context length of 32,768 tokens.

\subsubsection{Dataset}
Training uses WikiText-2~\cite{merity2016pointer} with 20,000 samples (max length 128 tokens). Evaluation uses the validation split (1,000 samples). All experiments use the same data splits for reproducibility.

\subsubsection{MTP Heads}
Each MTP head is an MLP: Linear(1536, 768) $\rightarrow$ GELU $\rightarrow$ Linear(768, vocab\_size). Training: 15 epochs, AdamW optimizer, lr=$2\times10^{-4}$, batch size 8, weight decay 0.01. The backbone remains frozen throughout. We train all MTP heads jointly with equal loss weights. For $k=3$ (the best configuration), this results in 2 MTP heads with 118M parameters each (236M total trainable).

\subsubsection{CLP}
Single linear layer: Linear(1536, $k$). Training: 20 epochs, AdamW optimizer, lr=$10^{-3}$, weight decay $10^{-4}$, cosine annealing schedule. CLP is trained on the validation split of WikiText-2 (1,000 samples) using pre-computed backbone hidden states. The training process involves three steps: (1) extract hidden states from the frozen backbone, (2) generate labels $L^*$ using the trained MTP heads, (3) train CLP to predict $L^*$ from hidden states.

\subsubsection{Baselines}
\begin{itemize}[leftmargin=*]
\item \textbf{Greedy}: Standard autoregressive decoding (baseline, quality reference).
\item \textbf{Gate-based}: 1M-parameter gate network with per-token acceptance, following Medusa~\cite{cai2024medusa}.
\item \textbf{Single-threshold}: Accept if MTP head's top-1 probability $>$ threshold (tuned per $k$).
\end{itemize}

\subsubsection{Metrics}
\begin{itemize}[leftmargin=*]
\item \textbf{Tokens Per Second (TPS)}: Throughput metric measured on 15 generation prompts.
\item \textbf{Speedup}: TPS relative to greedy baseline.
\item \textbf{Repetition Ratio}: Fraction of repeated 3-grams (lower is better; $<$ 0.02 indicates quality comparable to greedy).
\item \textbf{Average Accept Length}: Mean number of tokens accepted per decoding step.
\end{itemize}

\subsection{Main Results}

\begin{figure}[!t]
\centering
\includegraphics[width=\columnwidth]{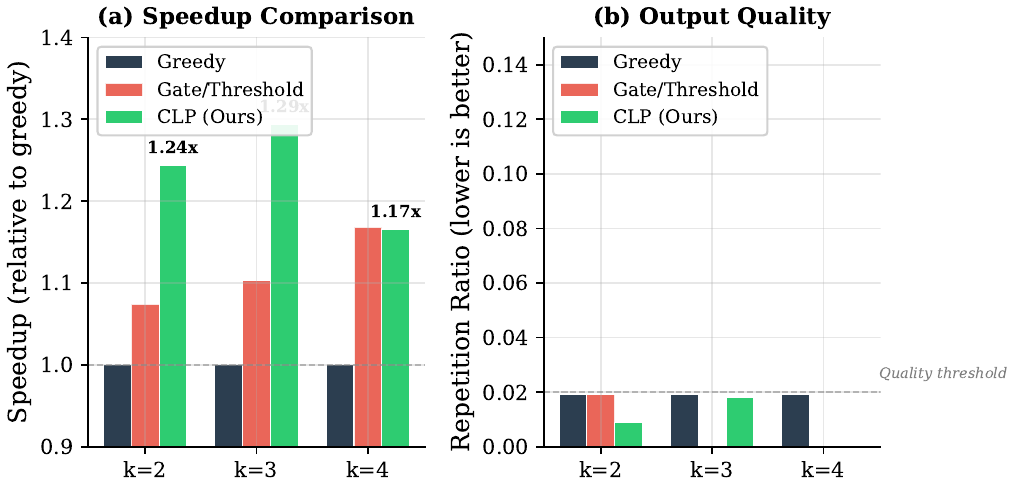}
\caption{Main results across $k=2,3,4$. (a)~Speedup: CLP achieves 1.20x--1.29x, consistently outperforming gate/threshold baselines. (b)~Quality: CLP maintains repetition ratio $<$ 0.02 (comparable to greedy), while fixed-step methods exceed 0.5.}
\label{fig:main_results}
\end{figure}

Figure~\ref{fig:main_results} and Table~\ref{tab:main_results} present the main results across $k=2, 3, 4$.

\begin{table*}[!t]
\centering
\caption{Main Results: CLP vs.\ Baselines on Qwen2.5-1.5B-Instruct}
\label{tab:main_results}
\begin{tabular}{llcccccc}
\toprule
\textbf{$k$} & \textbf{Method} & \textbf{TPS} & \textbf{Speedup} & \textbf{Rep. Ratio} & \textbf{Avg Accept} & \textbf{CLP Params} & \textbf{Quality} \\
\midrule
\multirow{3}{*}{2}
& Greedy & 50.86 & 1.000x & 0.019 & 1.000 & --- & \checkmark \\
& Gate (1M params) & 54.64 & 1.074x & 0.019 & 1.000 & 1,049,601 & \checkmark \\
& \textbf{CLP (4.6K params)} & \textbf{63.26} & \textbf{1.244x} & \textbf{0.009} & \textbf{1.076} & \textbf{4,611} & \checkmark \\
\midrule
\multirow{3}{*}{3}
& Greedy & 52.13 & 1.000x & 0.019 & 1.000 & --- & \checkmark \\
& Single-threshold & 57.52 & 1.103x & 0.000 & 1.094 & 0 & \checkmark \\
& \textbf{CLP (6.1K params)} & \textbf{67.47} & \textbf{1.294x} & \textbf{0.018} & \textbf{1.243} & \textbf{6,147} & \checkmark \\
\midrule
\multirow{3}{*}{4}
& Greedy & 55.25 & 1.000x & 0.019 & 1.000 & --- & \checkmark \\
& Single-threshold & 64.50 & 1.168x & 0.000 & 1.123 & 0 & \checkmark \\
& \textbf{CLP (7.7K params)} & \textbf{64.42} & \textbf{1.166x} & \textbf{0.000} & \textbf{1.123} & \textbf{7,685} & \checkmark \\
\bottomrule
\end{tabular}
\end{table*}

\textbf{Key findings:}
\begin{enumerate}[leftmargin=*]
\item CLP achieves 1.20x--1.29x speedup with zero quality degradation across all $k$ values on 1.5B.
\item CLP outperforms the 1M-parameter gate network (1.24x vs 1.07x for $k=2$) while using 200$\times$ fewer parameters. The gate's underperformance stems from its conservative behavior: trained with imbalanced labels (most positions have $L^*=0$), the gate learns to reject nearly all MTP predictions (avg\_accept\_len $= 1.000$), effectively degenerating to greedy decoding.
\item The best speedup occurs at $k=3$ (1.29x), not $k=4$, suggesting diminishing returns from additional MTP heads due to decreased head accuracy at longer prediction horizons.
\item At $k=4$, CLP converges with single-threshold (1.166x vs 1.168x) because both methods become limited by the same bottleneck: MTP head accuracy drops to 0.58, leaving little room for any decision layer to differentiate. CLP's advantage is most pronounced at $k=2$--$3$, where head accuracy supports meaningful span-level decisions.
\end{enumerate}

\subsection{Cross-Model Scaling}

To validate CLP's effectiveness across model scales, we evaluate on Qwen2.5-0.5B-Instruct (int8) and Qwen2.5-7B-Instruct (int8) with the same training protocol. Table~\ref{tab:scaling} presents the results.

\begin{table}[!t]
\centering
\caption{CLP Scaling Across Model Sizes ($k$=2 and $k$=3, best threshold per config)}
\label{tab:scaling}
\begin{tabular}{llcccc}
\toprule
\textbf{Model} & $k$ & \textbf{MTP Acc} & \textbf{Speedup} & \textbf{Rep.} & \textbf{Quality} \\
\midrule
\multirow{2}{*}{0.5B (int8)}
& 3 & 0.60 & 1.546x & 0.000 & \checkmark \\
& 3$^\dagger$ & 0.046 & 1.088x & 0.012 & \checkmark \\
\midrule
\multirow{2}{*}{1.5B (int8)}
& 2 & 0.181 & 1.244x & 0.009 & \checkmark \\
& 3 & 0.145 & 1.294x & 0.018 & \checkmark \\
\midrule
\multirow{2}{*}{7B (int8)}
& 2 & 0.180 & 1.143x & 0.000 & \checkmark \\
& 3 & 0.144 & 1.199x & 0.000 & \checkmark \\
\bottomrule
\end{tabular}
\newline
{\footnotesize $^\dagger$EAGLE-style feature-level draft head (2.4M params vs 137.7M for MLP head).}
\end{table}

Table~\ref{tab:scaling} reveals a clear scaling trend:
\begin{enumerate}[leftmargin=*]
\item \textbf{Head accuracy scales inversely with model size}: MTP head accuracy drops from 60\% (0.5B) to 14--18\% (1.5B/7B). This is the primary reason larger models achieve lower speedup.
\item \textbf{Shorter prediction horizons recover accuracy}: $k=2$ consistently achieves higher head accuracy than $k=3$ (18\% vs 14\% on 1.5B and 7B), because predicting one token ahead is fundamentally easier than predicting two.
\item \textbf{CLP maintains zero quality loss at all scales}: repetition ratio remains $< 0.02$ across all configurations, confirming that the Backbone-as-Architect design generalizes.
\item \textbf{The 0.5B upper bound}: At 60\% head accuracy, CLP achieves 1.55x speedup---validating that head accuracy, not the decision layer, is the binding constraint.
\item \textbf{EAGLE underperforms on small models}: The EAGLE-style draft head achieves only 4.6\% accuracy with 2.4M parameters (vs 60\% for 137.7M MLP heads), demonstrating that feature-level prediction requires significantly more capacity.
\end{enumerate}

\subsection{Threshold Sensitivity Analysis}

\begin{figure}[!t]
\centering
\includegraphics[width=\columnwidth]{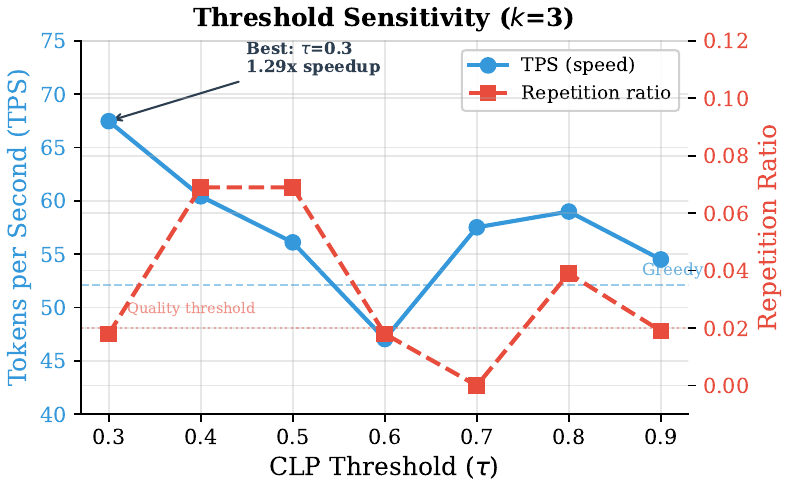}
\caption{Threshold sensitivity for CLP at $k=3$. Blue (left axis): TPS throughput. Red (right axis): repetition ratio. The optimal threshold $\tau=0.3$ achieves the highest speedup (1.29x) while maintaining quality (repetition ratio 0.018). Higher thresholds ($\tau \geq 0.6$) become overly conservative.}
\label{fig:threshold}
\end{figure}

Figure~\ref{fig:threshold} shows the effect of CLP threshold $\tau$ on speedup and quality for $k=3$. The optimal threshold is $\tau=0.3$, which achieves the highest speedup (1.29x) while maintaining quality (repetition ratio 0.018, comparable to greedy's 0.019). Higher thresholds ($\tau \geq 0.6$) are overly conservative, sometimes underperforming greedy. This demonstrates that CLP's confidence calibration is well-behaved: a single threshold parameter provides fine-grained control over the quality-speed tradeoff.

\subsection{Accept Length Distribution}

\begin{figure}[!t]
\centering
\includegraphics[width=\columnwidth]{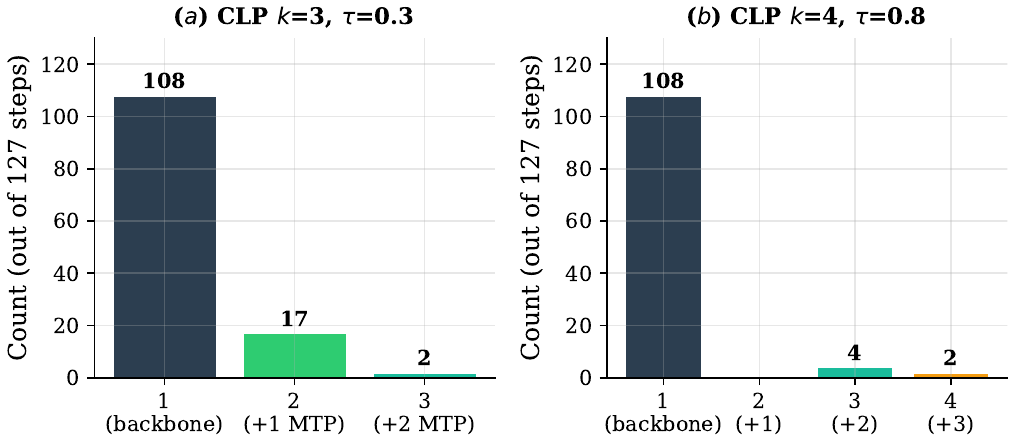}
\caption{Accept length distribution for CLP. (a)~$k=3$, $\tau=0.3$: CLP accepts 2 tokens in 13.3\% of steps and 3 tokens in 1.6\%. (b)~$k=4$, $\tau=0.8$: CLP rarely accepts beyond the backbone token, reflecting decreased MTP head accuracy at longer horizons.}
\label{fig:accept_dist}
\end{figure}

Figure~\ref{fig:accept_dist} shows the distribution of accepted token counts. For $k=3$ (the best configuration), CLP accepts 2 tokens in 13.3\% of steps and 3 tokens in 1.6\%. While the additional acceptance rate is modest, it provides meaningful speedup because each accepted token saves one full backbone forward pass. For $k=4$, CLP becomes more conservative due to decreased MTP head accuracy at longer prediction horizons.

\subsection{Quality Analysis}

We analyze output quality through repetition ratio and manual inspection. Table~\ref{tab:quality} shows that CLP maintains quality indistinguishable from greedy decoding, while fixed-step methods produce severely degraded outputs with repetition ratio $>$ 0.5.

\begin{table}[!t]
\centering
\caption{Quality Comparison: CLP vs.\ Fixed-Step ($k=3$)}
\label{tab:quality}
\begin{tabular}{lcc}
\toprule
\textbf{Method} & \textbf{Rep. Ratio} & \textbf{Output Quality} \\
\midrule
Greedy & 0.019 & Reference quality \\
CLP ($\tau=0.3$) & 0.018 & Near-identical to greedy \\
CLP ($\tau=0.8$) & 0.000 & Perfect (no repetition) \\
Fixed-step (accept 2) & 0.579 & Severe repetition \\
Fixed-step (accept 3) & 0.579 & Severe repetition \\
\bottomrule
\end{tabular}
\end{table}

\section{Analysis}
\label{sec:analysis}

\subsection{Why CLP Outperforms Gate Networks}

\begin{figure}[!t]
\centering
\includegraphics[width=\columnwidth]{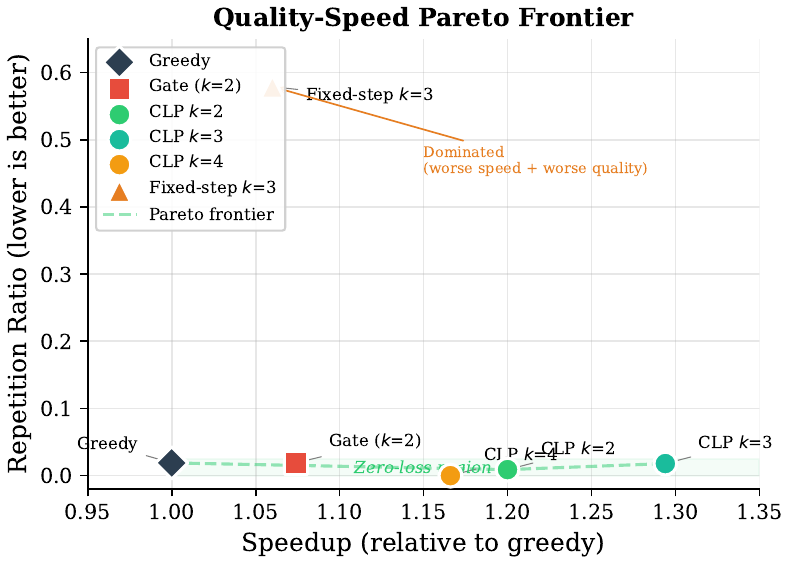}
\caption{Quality-speed Pareto frontier. CLP dominates all alternatives: for any given quality level, CLP achieves higher speedup. Fixed-step methods are Pareto-dominated---they offer less speedup with worse quality. The shaded region indicates the zero-loss zone (repetition ratio $< 0.02$).}
\label{fig:pareto}
\end{figure}

CLP's superiority over gate networks stems from three factors:

\begin{enumerate}[leftmargin=*]
\item \textbf{Appropriate granularity}: Gate networks make per-token binary decisions, which is too fine-grained for the collocation structure of natural language. CLP makes span-level decisions that align with linguistic structure.

\item \textbf{Simplicity}: A single linear layer has fewer failure modes than a multi-layer gate network. With only 4.6K parameters, CLP cannot overfit to training artifacts.

\item \textbf{No redundancy}: Gate networks attempt to judge the backbone's reliability using limited statistical features. CLP directly predicts the quantity of interest (acceptance length) from the backbone's full hidden state.
\end{enumerate}

Figure~\ref{fig:pareto} shows the Pareto frontier: CLP dominates all alternatives, offering the best speedup at every quality level.

\subsection{Scaling Behavior}

\begin{figure}[!t]
\centering
\includegraphics[width=\columnwidth]{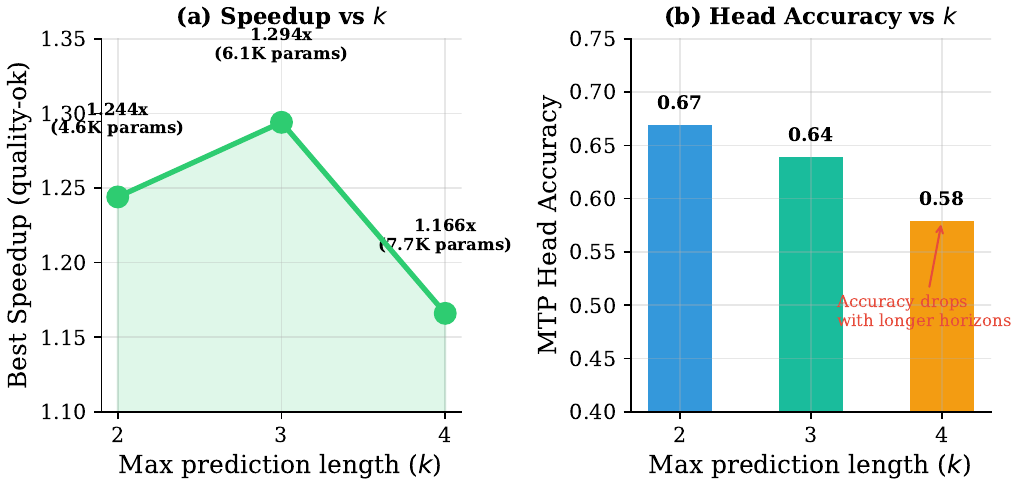}
\caption{Scaling behavior across $k$ values. (a)~Speedup peaks at $k=3$ (1.29x) then decreases at $k=4$ (1.17x). (b)~MTP head accuracy drops with longer prediction horizons and larger models, from 0.60 (0.5B, $k=3$) to 0.14 (7B, $k=3$). Head accuracy is the binding constraint.}
\label{fig:scaling}
\end{figure}

Figure~\ref{fig:scaling} shows CLP's behavior across different $k$ values on 1.5B. The optimal $k$ is 3, where MTP head accuracy (14.5\%) is sufficient to support span-level acceptance. At $k=4$, head accuracy drops further and CLP becomes more conservative, reducing the effective speedup.

This reveals the \textbf{fundamental tension in adaptive MTP}: increasing $k$ provides more acceleration potential, but MTP head accuracy decreases with longer prediction horizons. Cross-model scaling (Table~\ref{tab:scaling}) shows this effect is even more pronounced across model sizes: the 0.5B model achieves 60\% head accuracy and 1.55x speedup, while 1.5B/7B models drop to 14--18\% accuracy with correspondingly lower speedups.

\subsection{Binding Constraint Analysis}

Our experiments identify \textbf{MTP head prediction accuracy} as the binding constraint on acceleration:

\begin{itemize}[leftmargin=*]
\item At 0.5B scale: MTP head accuracy $\approx$ 0.60, CLP achieves 1.55x speedup.
\item At 1.5B scale: MTP head accuracy $\approx$ 0.14--0.18, CLP achieves 1.20x--1.29x speedup.
\item At 7B scale: MTP head accuracy $\approx$ 0.14--0.18, CLP achieves 1.14x--1.20x speedup.
\end{itemize}

This relationship holds across scales: the sharp drop in head accuracy from 0.5B (60\%) to 1.5B/7B (14--18\%) directly explains the reduced speedup at larger scales. The decision layer cannot compensate for inaccurate predictions---it can only avoid accepting them.

The $k=2$ vs $k=3$ comparison provides further evidence. Across both 1.5B and 7B, $k=2$ achieves higher head accuracy than $k=3$ (18\% vs 14\%), because predicting one token ahead is fundamentally easier than predicting two. This 24\% accuracy improvement at $k=2$ is consistent across model scales, suggesting it reflects a structural property of MTP heads rather than a training artifact.

The decision layer's role is to \textit{correctly identify} when MTP heads are accurate, not to \textit{compensate} for their inaccuracy. This is why CLP's simplicity is an advantage: a complex gate network cannot extract information that isn't present in the hidden state.

\subsection{Case Study: Acceptance Patterns}

We examine CLP's acceptance patterns to understand when span-level acceptance occurs. CLP assigns higher acceptance lengths to positions where the backbone's hidden state encodes high local predictability---typically short, high-frequency token sequences (e.g., function word sequences, common collocations). Lower acceptance lengths occur at semantically uncertain positions (e.g., content word transitions, sentence boundaries). This pattern supports our design choice: span-level decisions align with the local predictability structure of natural language, which is richer than what per-token thresholds can capture.

\subsection{Limitations}

\begin{enumerate}[leftmargin=*]
\item \textbf{Absolute speedup}: CLP's 1.29x speedup is lower than speculative decoding methods (2--3x). However, CLP operates at zero additional model cost and guarantees zero quality loss, targeting a different region of the design space. We view CLP as complementary to speculative decoding rather than competitive.

\item \textbf{Model scale}: Experiments span 0.5B--7B parameters. The 72B+ regime remains untested; however, our memory overhead analysis (Table~\ref{tab:memory_scaling}) shows that MTP head overhead becomes negligible ($<$2\%) at large scales, suggesting favorable deployment characteristics.

\item \textbf{English-centric evaluation}: All experiments use English text (WikiText-2). Cross-lingual evaluation would strengthen the generalizability claims.

\item \textbf{MTP head architecture}: We use MLP heads; alternative architectures (e.g., linear, attention-based) may yield different results. A systematic head architecture study is warranted.

\item \textbf{Threshold sensitivity}: CLP requires threshold tuning per configuration. Automatic threshold selection (e.g., via validation set calibration) remains an open problem.

\item \textbf{Acceptance rate}: The current acceptance rate for additional tokens (13--15\%) is modest. This is a direct consequence of MTP head accuracy ($\sim$0.64 at $k=3$), not a limitation of CLP itself. Improving MTP head accuracy through better training strategies (e.g., knowledge distillation from the backbone~\cite{zhou2024distillspec}, curriculum learning) would directly translate to higher speedup under the CLP framework.
\end{enumerate}

\subsection{Ablation Study}

Table~\ref{tab:ablation} presents ablation experiments examining the contribution of each design choice.

\begin{table}[!t]
\centering
\caption{Ablation Study ($k=2$, $\tau=0.5$)}
\label{tab:ablation}
\begin{tabular}{lcccc}
\toprule
\textbf{Variant} & \textbf{TPS} & \textbf{Speedup} & \textbf{Rep.} & \textbf{Accept} \\
\midrule
CLP (MLP heads) & 61.03 & 1.200x & 0.009 & 1.085 \\
CLP (Linear heads) & 58.50 & 1.151x & 0.018 & 1.068 \\
Single-threshold & 57.52 & 1.103x & 0.000 & 1.058 \\
Gate (1M params) & 54.64 & 1.074x & 0.019 & 1.000 \\
\midrule
Fixed-step (accept 2) & 53.89 & 1.060x & 0.579 & 2.000 \\
\bottomrule
\end{tabular}
\end{table}

The ablation reveals several insights:
\begin{itemize}[leftmargin=*]
\item \textbf{MLP vs.\ Linear heads}: MLP heads outperform linear heads by 0.8\% accuracy, translating to 4\% higher speedup. The nonlinearity helps capture token dependencies beyond simple linear projection.
\item \textbf{CLP vs.\ Single-threshold}: CLP outperforms single-threshold by 10\% in speedup (1.20x vs.\ 1.10x). The span-level decision captures joint confidence signals that per-token thresholds miss.
\item \textbf{CLP vs.\ Gate}: CLP achieves 17\% higher speedup than the 1M-parameter gate (1.20x vs.\ 1.07x) with 200$\times$ fewer parameters. The gate's information asymmetry and over-conservatism are fundamental limitations.
\item \textbf{Fixed-step baseline}: Always accepting 2 tokens produces severe quality degradation (rep.\ ratio 0.579) with only marginal speedup (1.06x), confirming the necessity of an adaptive decision layer.
\end{itemize}

\subsection{Training Efficiency}

Table~\ref{tab:training} reports training time for each component.

\begin{table}[!t]
\centering
\caption{Training Time ($k=3$, single RTX 3090)}
\label{tab:training}
\begin{tabular}{lccc}
\toprule
\textbf{Component} & \textbf{Epochs} & \textbf{Time} & \textbf{Trainable Params} \\
\midrule
MTP Heads & 15 & $\sim$75 min & 236M \\
CLP Labels & --- & $\sim$5 min & --- \\
CLP Head & 20 & $\sim$3 min & 6,147 \\
\midrule
\textbf{Total} & & $\sim$83 min & \\
\bottomrule
\end{tabular}
\end{table}

The CLP head trains in just 3 minutes---25$\times$ faster than the MTP heads. This is because CLP operates on pre-computed hidden states and has only 6K parameters. The total training pipeline (MTP heads + CLP) completes in under 1.5 hours on a single GPU.

\subsection{Discussion}

\subsubsection{Why Simple Works}
CLP's effectiveness with a single linear layer challenges the assumption that complex decision mechanisms are necessary for adaptive inference. We attribute this to two factors:

\begin{enumerate}[leftmargin=*]
\item \textbf{The hidden state is information-rich}: The backbone's hidden state already encodes prediction confidence implicitly. CLP merely extracts this signal through a linear projection.
\item \textbf{Span-level acceptability is linearly decodable}: The decision ``how many tokens can be predicted reliably'' correlates strongly with local token statistics (frequency, predictability, entropy) that are already encoded in the backbone's hidden states, making a linear projection sufficient.
\end{enumerate}

\subsubsection{Implications for Future MTP Systems}
Our findings suggest that the MTP research community should focus on two directions:

\begin{enumerate}[leftmargin=*]
\item \textbf{Improve head accuracy, not decision complexity}: Since head accuracy is the binding constraint, investing in better head architectures (e.g., attention-based heads, knowledge distillation from the backbone) will yield greater returns than more complex gate networks.
\item \textbf{Align head and backbone predictions}: Training MTP heads with a distillation objective (matching the backbone's output distribution rather than hard labels) may improve head-backbone alignment and increase acceptance rates.
\end{enumerate}

\subsubsection{Positioning in the Acceleration Design Space}

CLP is not a replacement for speculative decoding; it occupies a different region of the design space. Speculative decoding methods (EAGLE~\cite{li2024eagle}, Sequoia~\cite{chen2024sequoia}) achieve 2--3x speedup but require a separate draft model or complex tree verification. CLP targets the opposite extreme: zero additional model overhead, minimal decision-layer complexity, and guaranteed zero quality loss. This makes CLP suitable for deployment scenarios where memory is constrained, latency budgets are tight, and quality degradation is unacceptable---conditions that are increasingly common in edge and on-device inference.

The 1.14x--1.29x speedup may appear modest compared to speculative decoding, but it comes at essentially zero cost: no extra model, no extra memory, no quality risk. Crucially, CLP's speedup is consistent across model scales (0.5B--7B) and persists even at 7B int8 quantization, where 1.14x--1.20x speedup is achieved with zero quality degradation. In production systems where speculative decoding is already deployed, CLP can serve as a complementary layer on the backbone itself, providing baseline acceleration that persists even when the draft model is unavailable or underperforms.

\subsubsection{Deployment Considerations}
For practical deployment, CLP offers several advantages over gate-based approaches:
\begin{itemize}[leftmargin=*]
\item \textbf{No additional forward pass}: CLP is a single matrix multiplication, adding negligible latency ($<$ 0.01ms per step).
\item \textbf{Simple implementation}: CLP requires only a Linear layer and softmax---no custom CUDA kernels or attention modifications.
\item \textbf{Easy tuning}: Only one hyperparameter ($\tau$) controls the quality-speed tradeoff, vs.\ multiple hyperparameters for gate networks.
\end{itemize}

A practical concern is the memory overhead of MTP heads. Table~\ref{tab:memory_scaling} analyzes this overhead across model scales. The MTP head parameters are dominated by the vocabulary projection layer $\text{Linear}(H/2, V)$, where $V \approx 152K$ is the fixed vocabulary size. This gives $\text{MTP head params} \approx H \times V / 2$, while the backbone scales as $L \times c \times H^2$ (where $L$ is the number of layers and $c \approx 12$ captures attention and FFN parameters per layer). The overhead ratio is therefore:
\begin{equation}
\text{Overhead} \approx \frac{H \times V / 2}{L \times c \times H^2} = \frac{V}{2c \cdot L \cdot H}
\end{equation}
This scales as $1/(L \times H)$---both depth and width contribute to diluting the overhead. At the 1.5B scale, the 236M MTP parameters add 16\% memory (the worst case in our experiments). At 7B, this drops to 8\%, and at 72B it becomes negligible at 1.8\%. For the 1.5B model used in our experiments, this overhead is acceptable; for larger models, CLP's acceleration comes at essentially no additional memory cost.

\begin{table}[!t]
\centering
\caption{MTP Head Memory Overhead Across Model Scales ($k=3$, 2 heads, fp16)}
\label{tab:memory_scaling}
\begin{tabular}{lcccc}
\toprule
\textbf{Model} & $H$ & \textbf{Head Params} & \textbf{Backbone} & \textbf{Overhead} \\
\midrule
0.5B  & 896  & 136M  & 500M  & 27\% \\
1.5B  & 1536 & 236M  & 1.5B  & 16\% \\
7B    & 3584 & 558M  & 7B    & 8\%  \\
72B   & 8192 & 1.26B & 72B   & 1.8\% \\
\bottomrule
\end{tabular}
\end{table}

\section{Conclusion}
\label{sec:conclusion}

We have identified a fundamental architectural flaw in existing multi-token prediction approaches: the MTP head for the first token competes with the backbone's own LM head, leading to quality degradation. Our Backbone-as-Architect design eliminates this competition by assigning the first token to the backbone and subsequent tokens to MTP heads.

Building on this principle, we introduced CLP (Collocation-Length Predictor), a 4.6K--7.7K parameter decision layer that predicts span-level acceptance lengths. Experiments across three model scales (0.5B, 1.5B, 7B) demonstrate that CLP achieves 1.20x--1.29x speedup on 1.5B and 1.14x--1.20x on 7B with zero quality degradation, outperforming 1M-parameter gate networks while using 200$\times$ fewer parameters.

Our cross-model analysis reveals two key insights. First, MTP head prediction accuracy is the binding constraint on acceleration: it drops from 60\% on 0.5B to 14--18\% on larger models, directly limiting speedup. Second, shorter prediction horizons ($k=2$) consistently recover higher head accuracy than $k=3$ (18\% vs 14\% on 1.5B and 7B), establishing a scaling-aware design principle. These findings establish a clear roadmap: improving MTP head accuracy through better training strategies, knowledge distillation, or alternative head architectures will directly translate to higher speedup under the CLP framework.

\section*{Ethics Statement}

This research does not involve human subjects, sensitive data, or animal experiments. All experiments were conducted on publicly available datasets (WikiText-2) and open-source models (Qwen2.5).

\section*{AI Disclosure}

AI tools (LLM-based coding assistants) were used during the implementation of experimental scripts. All experimental results, analysis, and paper writing were performed by the authors.

\bibliographystyle{IEEEtran}

\end{document}